# Exploring Sentiment Analysis Techniques in Natural Language Processing: A Comprehensive Review

**Karthick Prasad Gunasekaran[1]**

College of Information and Computer Sciences, University of Massachusetts, Amherst, United States[1]

*Abstract*

Sentiment analysis (SA) is the automated process of detecting and understanding the emotions conveyed through written text. Over the past decade, SA has gained significant popularity in the field of Natural Language Processing (NLP). With the widespread use of social media and online platforms, SA has become crucial for companies to gather customer feedback and shape their marketing strategies. Additionally, researchers rely on SA to analyze public sentiment on various topics. In this particular research study, a comprehensive survey was conducted to explore the latest trends and techniques in SA. The survey encompassed a wide range of methods, including lexicon-based, graph-based, network-based, machine learning, deep learning, ensemble-based, rule-based, and hybrid techniques. The paper also addresses the challenges and opportunities in SA, such as dealing with sarcasm and irony, analyzing multi-lingual data, and addressing ethical concerns. To provide a practical case study, Twitter was chosen as one of the largest online social media platforms. Furthermore, the researchers shed light on the diverse application areas of SA, including social media, healthcare, marketing, finance, and politics. The paper also presents a comparative and comprehensive analysis of existing trends and techniques, datasets, and evaluation metrics. The ultimate goal is to offer researchers and practitioners a systematic review of SA techniques, identify existing gaps, and suggest possible improvements. This study aims to enhance the efficiency and accuracy of SA processes, leading to smoother and error-free outcomes.

## I. INTRODUCTION

Sentiment analysis is the process of recognizing and extracting subjective information from textual data. It includes analyzing opinions, attitudes, emotions, and feelings articulated in a text and categorizing them as positive, negative, or neutral sentences [1]. SA has gained a lot of popularity in recent years due to the abundance of user-generated content on online platforms, which includes social media, blogs, reviews, and discussion forums. Businesses use this sentiment analysis to learn about customer feedback, monitor brand reputation, and make strategic decisions accordingly. This survey reviews the top existing research techniques for SA in NLP [2]. Researchers have discussed the most recent strategies for sentiment analysis in NLP in this survey study. An overview of the many approaches used for sentiment analysis, including deep learning techniques [3], rule-based methods [4], machine learning techniques, sentiment strength detection methods [5], swarm intelligence based methods, sentiment lexicon expansion methods, Bayesian methods, and pattern-based methods, is reviewed. Difficulties and limits of SA approaches [6], such as the absence of context, the existence of sarcasm and irony, have also been addressed in this research. The difficulty of dealing with multilingual and domain-specific data. Furthermore, researchers demonstrate SA applications in a variety of disciplines, including marketing, finance, politics, healthcare, and social media. The report concludes with a discussion of future research areas and remaining issues in sentiment analysis.

 This study begins with the fundamental ideas of SA, such as sentiment classification, feature extraction, and sentiment lexicons. The numerous forms of sentiment analysis are then discussed, including document-level, sentence-level, and aspect-based sentiment analysis. The survey compares rule-based, machine learning, and deep learning techniques for SA and explores their merits and demerits. Research also focuses on the difficulties of dealing with multilingual and cross-cultural sentiment analysis, as well as the ethical and privacy problems that come with sentiment analysis. This survey paper presents a detailed and state-of-the-art assessment of SA in NLP. This paper intends to be a valuable asset for NLP practitioners, researchers, and scholars.

*I.I SENTIMENT ANALYSIS:*

Sentiment analysis is the technique in NLP for extracting and fetching subjective information out of textual data. It also includes the categorization of text into negative, neutral or positive sentences. It involves deciphering the underlying sentiment of thoughts, attitudes, emotions, and sentiments portrayed in any natural language. SA involves deciphering the underlying sentiment of thoughts, attitudes, emotions, and sentiments portrayed in any natural language.

Sentiment analysis may be used on a variety of textual data sources, including social media poles, buyer reviews, news stories, and product descriptions.

Figure 1 explains the general framework of SA process where the first step is to input the text data, secondly there are three steps for data pre-processing which includes tokenization, stopword filtering and stemming of the textual data. After that the most crucial part comes which is the choice of the classification method.

The purpose of SA is to comprehend both the entire sentiment of a textual information and the sentiment regarding individual features or entities referenced in the text. The terms feeling, view, opinion, and belief are used interchangeably, there are distinctions between them. Opinion is the conclusion which is subject to dispute (because of diverse expert perspectives), view is a personal opinion by any individual, belief refers to conscious acceptance and intellectual assent whereas sentiment is a view that expresses one's sentiments.

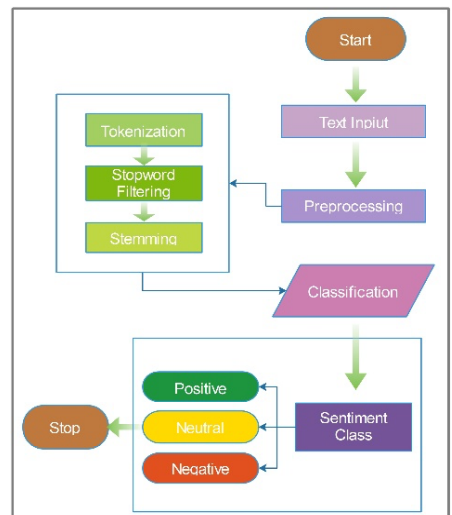

Figure 1: General framework of SA process.

Sentiment analysis may be used on a variety of textual data sources, including social media poles, buyer reviews, news stories, and product descriptions.

*I.II SENTIMENT ANALYSIS APPLICATIONS:*

Sentiment analysis has emerged to be an important field to explore in NLP due to its wide range of applications and ability to reveal significant insights from unstructured textual data [7]. Sentiment analysis has been applied in opinion mining, brand monitoring, market research, customer service, political analysis, and many more fields. The rise of social media platforms has made sentiment analysis even more essential, as individuals share their ideas and experiences on a massive scale.

Opinion analysis is also used in healthcare to analyse patient feedback, education to evaluate student opinion, and finance to forecast market trends. SA has variety of applications in real world including social media monitoring, product development, analysis of political situations, education, finance and e-commerce. It is also applicable in healthcare sector in gauging patients' feedback and identifying the areas of improvement. SA may be used in financial services to analyse customer reviews and discover problems with financial possessions and services. It may also be used to analyse finance-related social media conversations and determine public opinion towards financial institutions and legislation. It may be used in e-commerce in order to detect producer service flaws and enhance customer experience. Figure 2 represents some real-world applications of SA.

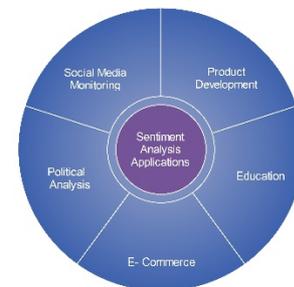

Figure 2: Real world applications of Sentiment Analysis.

Sentiment analysis may be used in education to examine student remarks and find areas for improvement in educational methodologies. It may also be used to analyse twitter talks regarding educational subjects in order to determine public opinion about educational policies and institutions. In sports, SA may be used to examine public opinion towards sports teams, individuals, and events. This may help sports organisations and teams enhance their marketing strategies and more effectively communicate with fans.

*I.III SENTIMENT ANALYSIS APPROACHES:*

SA is the branch of NLP and machine learning (ML) which includes analysing text using various approaches and algorithms. Rule-based methods, ML algorithms including Naive Bayes and Support Vector Machines (SVM), and deep learning models such as Convolutional Neural Networks (CNN) and Recurrent Neural Networks (RNN) are examples of these techniques. Twitter, has emerged as a popular medium for sentiment research due to its real-time nature and large user community, it has over 330 million active users across the globe which makes it a valuable source of data for SA researchers. The researchers have reviewed a variety of methodologies and techniques for sentiment analysis on Twitter, such as lexicon-based methods, deep learning models, and machine learn (ML) algorithms. However, due to social media's quick expansion and innovation, there are several obstacles and possibilities related with sentiment analysis on Twitter data. This study has compared two most commonly used approaches for SA on twitter data. These methods are
1. Lexicon-Based Approach
2. Machine Learning (ML)Approach

Detailed analysis of above-mentioned approaches is being formed during the study. Process, training time, computational complexity, evaluation measures and accuracy have been compared carefully. The study also focuses on the detailed assessment of the datasets available for SA. Figure 3 showcases subcategories of two most widely SA approaches.

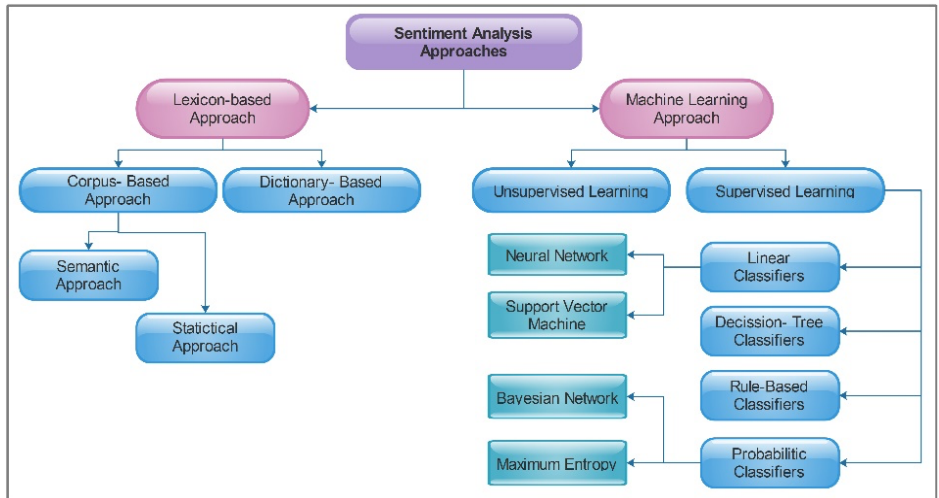

Figure 3: Widely used Sentiment analysis approaches.

In recent years, Twitter sentiment analysis has been the topic of substantial research, with multiple publications suggesting various methodologies and ways to increase its accuracy. With research concentrating on feature selection, domain adaption, and aspect-based sentiment analysis, the employment of ML algorithms, lexicon-based techniques, and deep learning models has grown in popularity. They are widely used in recent researches on SA to measure the behaviour of the common people.

*I.III.I) LEXICON-BASED APPROACH*

A lexicon-based method to sentiment analysis includes analysing the sentiment of a piece of text using a pre-defined list of terms with associated sentiment scores [8]. Typically, the sentiment score is a numerical number that measures the strength of the feeling, such as a positive score for positive statement and a negative score for negative statement. The vocabulary can be built manually or automatically using machine learning techniques. Typically, the lexicon is created by picking a collection of words that are known to indicate positive or negative sentiment and assigning a score to each word depending on the intensity of sentiment [9]. The sentiment score for a piece of text is determined using lexicon-based sentiment analysis by adding the sentiment scores [10]. Lexicon-based techniques is easier to deploy and requiring less training data than machine learning approaches. They may not be as accurate as machine learning algorithms, because they do not consider the context of the words or the links between words in a phrase. These techniques are mainly used in social media monitoring and political analysis applications.

*I.III.II) MACHINE LEARNING APPROACH*

ML approaches have gain popularity for semantic analysis because they can learn from vast volumes of data without being explicitly trained [11]. Supervised learning, unsupervised learning, transfer learning and reinforcement learning are few sub categories of ML approaches. There are some examples of typical ML approaches for semantic sentiment analysis:

i. Supervised learning: This is the training of a model using a labelled dataset in which the input text is linked with its associated sentiment label [12]. The model learns to recognise textual patterns associated with various sentiment labels and can subsequently categorise new, unlabelled material accordingly.
ii. Unsupervised learning: The model is not provided labelled input and must recognise patterns and structures in the data on its own. This may be used for activities like grouping related papers or themes.

iii. Deep learning: Here the models are multi-layered neural networks that can learn complicated patterns and correlations in data. They've been utilised to do tasks including sentiment analysis, natural language processing, and speech recognition.

## II. COMPARATIVE ANALYSIS

Twitter is vastly used social media platforms where people from all walks of life share their opinion on various topics. There are numerous researches being performed on SA using twitter information exchange. As twitter a vast range of provides point of views from people across the world. In this section, a comprehensive comparison is being performed on the past researches. Data available on social networking websites is one of the most imperative inputs for projecting automobile sales. Furthermore, other factors like stock market valuations, have an impact on automobile purchasing power. To estimate monthly total car sales, Pai et al. [13] used multivariate regression models incorporating both data from social media and stock market standards and time series models. While dealing with multivariate regression data, least squares support vector regression (LSSVR) models were applied and three data types including sentiment scores, stock market values and hybrid data are used to predict the market trends and sales of vehicles in united states of America. Furthermore, seasonal factors were also taken care of as they also play an important role in vehicle sales. Results show that LSSVR models employing hybrid data with deseasonalizing processes can produce precise results than other models using different dataset. Liao et al. [14] presented an SA method that is capable of predicting user satisfaction about a product, views of public on current scenarios and their sentiments on joyous or disastrous situations. The main rationale for using CNN in image analysis and classification is that it can extract an area of features from global information and examine the link between these characteristics. It can improve analysis and classification accuracy. For NLP, text data components may be retrieved piece by piece and the connection between these aspects considered, but without considering context or the entire phrase, the emotion may be misinterpreted. Currently, CNN is one of the most successful approaches for picture classification; It contains a convolutional layer to dig information from a bigger piece of text, thus the technique employs a convolutional neural network for SA. A basic CNN is created and tested on pre-set standards. The results exhibit a remarkable performance in comparison with the existing techniques such as Naïve Bayes and SVM methods.

Stock market is one of the common application areas of SA. In the recent times social media specially, twitter is impeccable representative of public sentiment and opinion about current scenarios. Pagolu et al.[15] presented a paradigm to monitor stock market trends and predict the upcoming fluctuations using public sentiments on twitter. The study observed how well the changes in stock prices of an organization, the ups and downfalls are linked with the public opinions being expressed in tweets about that organization. The objective is to understand the authors' opinion from a piece of text. Two textual representation i-e Word2vec and Ngram are used for SA. The research applied SA and supervised learning framework to assess correlation between a company's stock trends as well as their opinions in tweets published by them. Additionally, favourable talks and tweets about a company on twitter or other social sites can undoubtably motivate individuals to invest in the company's stocks. This results in an increase in the company's reputation and financial standing which ultimately increases the stocks. Kang et al. [16] presented a Naïve Bayes(NB) SA architecture for analysis of twitter posts sharing. The architecture aimed to analyse the restaurant reviews using the Senti-Lexicon approach. As the previous methods used for this purpose do not provide the adequate insight of the customer sentiments. In this study, restaurant review was categorized into two main categories, positive and negative using improved NB approach. The improved NB method decreased the gap between positive & negative predictions to 3.6%. Moreover, when compared to SVM, the accuracy of this algorithm based on the senti-lexicon improved by 10.2% in recall and of 26.2% in precision, and it is improved by 5.6% in recall 1.9% in precision when Naive Bayes approach was applied.

Kouloumpis et al. [17] presented a research study on sentiment analysis of twitter messages using linguistic features. The study evaluated the expediency of present lexical resources and features that gather information about creative as well as informal language used in twitter messages. A supervised learning approach is used for solving the problem and it turned out to be very useful in terms of accuracy. The researchers also catered hashtags in order to build the training data. A paradigm of extracting twitter sentiments is presented by Agarwal et al. [18] in a research article where they explored individual sentiments of people on twitter regarding different topics by using data mining approaches including ML and SA. Part of speech (POS) prior-polarity features are used to categorize the sentiments into positive, negative ana neutral ones. Secondly, A tree kernel is employed to eliminate the requirement for time-consuming feature engineering. The novel features (when combined with previously suggested features) and the tree kernel performed in the same manner with minor differences in the outcomes. The proposed technique outdoing the state-of-the-art approaches, shown by extensive experimentations and analysis. Qiu et al. [19] have suggested a new advertising method called Dissatisfaction-oriented Advertising based on Sentiment Analysis (DASA) to solve the problem of traditional context based advertisement or correlating advertisements on web pages. According to the researchers, DASA will increase ad relevancy and user experience side by side. The algorithm used a rule-based method for extracting subject terms from opinion sentences linked with negative emotion, which are used as advertising keywords. A sample system for submitting product details for ad choice is also presented in the study. The trial findings on promotional keyword extraction and ad choice have shown that the suggested technique is successful. Neviarouskaya et al. [20] aimed to develop a framework to classify text using fine-grained attitudedescriptors.The created system is based on the complexity principle and a unique methodology based on semantically separate verb classes. This technique performed well on 1000 phrases describing individual experiences, with an average accuracy of 62% on the finegrained level (14 labels), 71% on the intermediate level (7 labels), and 80% on the top level (3 labels). Table 1 represents performance of different techniques used previously for SA. It also includes the datasets and algorithms being used in chronological order.

Table 1: Performance comparison of various Sentiment Analysis on Twitter data in previous decade.

| Year | Dataset | Model | Performance |
|---|---|---|---|
| 2018 | Yahoo finance data, Twitter dataset and hybrid data | LLSVR [13] | Hybrid data can improve accuracy. |
| 2017 | MR training STS Gold dataset | CNN [14] | 74.5% 68% |
| 2016 | Tweets collected using twitter API | Logistic Regression Lib SVM [15] | 69.01% 71.82% |
| 2012 | Twitter dataset | Improved Naive bayes Senti-Lexicon method [16] | 3.6% improved 26.2% improved |
| 2011 | EMOT HASH ISIEVE | Supervised learning approach [17] | 2 classes (+ve, -ve) +ve, -ve, Neutral +ve, -ve, Neutral |
| 2011 | Twitter dataset | Tree kernel model [18] | 64% |
| 2010 | Real-time twitter messages | DASA (Rule-based approach) [19] | 55% |
| 2010 | Fine-grained Middle-level Top-level | Lexicon-based semantic approach [20] | 62% 71% 88% |

### III. DISCUSSIONS AND FINDINGS

As discussed in the previous section, multiple techniques have been used to analyse sentiments in the posts published on twitter. Various application areas being stock exchange, monitoring sales of vehicles, restaurant reviews, social and political reviews and healthcare. Twitter data is often noisy and unstructured, data preparation is a critical step in sentiment analysis. Text normalization, stemming, and stop word removal are all techniques that can assist enhance the accuracy of sentiment analysis models. The research shows that the choice of the algorithm depends on the type of application it is being used for. Different techniques behave differently on the data and have different performance accordingly. This study focuses on the various ML algorithms used in sentiment analysis and opinion mining. SA integrated with ML techniques can be in useful for forecasting product appraisals and customer attitudes towards freshly introduced products. This study provides a detailed examination of several machine learning approaches and then compares their accuracy, benefits, and limits. While compared with unsupervised learning approaches, the supervised learning technique obtained 85% accuracy. Many more studies may be conducted, for instance, using the word2vec tool [21], a multilayer convolutional neural network, a bigger training datasets, and additional scenario or state assessments. Deep learning algorithms are sophisticated enough to address CV or speech recognition challenges, and CNN is the best choice for image analysis and classification [14]. SA, often known as opinion mining, is the art of automating identifying and extracting subjective information from text, such as attitudes, viewpoints, and emotions [18]. By assessing the mood of social media posts, businesses, governments, and individuals may gain exclusive insights into how people feel regarding their products, policies, and ideas. A tree kernel [18] is used to avoid to the requirement for time taking feature engineering in SA process. The tree kernel and novel features when paired together, performed really well. The extensive experimentation was performed to validate the finding. Another research study on SA shows that forecasting monthly car sales requires the data both, from twitter as well as stock market [13]. This will help to accurately predict the sales according to the market trends. The results of SA research can be used in a variety of fields, including marketing, politics, healthcare, and customer service. The results of the experimental process tells that there is higher association between rise or fall of stocks and word of mouth. Businesses, for example, might use sentiment analysis to enhance their products and services by analysing client comments on social media and online review sites. attitude analysis may be used in politics to measure public opinion and attitude towards political candidates or policies.

## IV. CONCLUSION

Availability of data on digital platforms and the need to analyse it for various purposes has gained a lot of popularity. Sentiment analysis is widely applicable in different industries. This study focused on SA using the sentiments showcased on twitter platform from various fields of life. The study mainly focused on ML techniques and Lexicon based methods used for SA on twitter data. The datasets that were analyzed are EMOT, HASH, ISIEVE, Twitter dataset, and some real time data as well. The survey discusses use cases, benefits, and limitations of sentiment analysis, as well as examples of effective employment of techniques. Study also talks about the limitations and constraints of using sentiment analysis in certain sectors like political analysis and healthcare, where the results might be difficult. It can be concluded that when SA is used rightly, it gives useful information of public opinion and which can result in making better business decisions. It also helps in predicting market trends. It is also concluded that the choice of keyword for twitter has a substantial impact on tweet search results and prediction accuracy. Twitter sentiment analysis can also be used for collecting geographical information and explore the unexplored areas. Furthermore, Lexicon based approaches can be utilized to assess public sentiments on ongoing situations whereas ML approaches can be applied to investigate market trends.